# LOSSLESS FITNESS INHERITANCE IN GENETIC ALGORITHMS FOR DECISION TREES

Last updated: March 6, 2009

Dimitris KALLES[1], Athanasios PAPAGELIS[2]

## ABSTRACT

When genetic algorithms are used to evolve decision trees, key tree quality parameters can be recursively computed and re-used across generations of partially similar decision trees. Simply storing instance indices at leaves is enough for fitness to be piecewise computed in a lossless fashion. We show the derivation of the (substantial) expected speed-up on two bounding case problems and trace the attractive property of lossless fitness inheritance to the divide-and-conquer nature of decision trees. The theoretical results are supported by experimental evidence.

## KEYWORDS

Decision trees, genetic algorithms, fitness inheritance, fitness approximation, learning speed-up.

## 1. Introduction

Decision trees have two key merits when compared to other concept learners. First, greedy algorithms have been shown to produce good results for an array of interesting problems and they requite relatively little computational power for the creation of the model of the underlying hypothesis; moreover, the representation of the model does not demand excessive memory. Second, by providing classifications and predictions that can be argued about, they advance our insight in the problem domain.

Advances in decision trees have been indicative of the introduction of sophisticated methodologies into the mainstream of research, industry and education. The really early approaches focused on optimizing the building of decision trees based on dynamic programming (Meisel & Michalopoulos, 1973). An NP-completeness result (Hyafil & Rivest, 1976) resulted in nearly a decade of silence, which was broken by the systematic use of heuristics for top-down decision tree induction (Breiman *et al.*, 1984; Quinlan, 1986). The abundant research in the field generated heuristics for nearly all stages of the decision tree life-cycle, theoretical results for optimal construction (Naumov, 1991), commercial software (Quinlan, 1993) and the field became ripe for textbook treatment (Mitchell, 1997; Rokach and Maimon, 2008).

As genetic algorithms started being increasingly employed into the investigation of NP-complete problems (Mitchell, 1996), some attempts were also made at trying to apply them to the building of decision trees. The earliest approach to using genetic programming in decision trees is probably traced back to Koza (1991) who first pointed out the suitability of the tree genome for decision tree building. Turney (1995) applied genetic algorithms to search through the space of decision trees generated by top-down learners (focusing on cost-sensitive learning). Later on Nikolaev & Slavov (1998) analyzed a global fitness landscape structure and its application on decision tree building, while Bot & Longdon (2000) used genetic programming to evolve linear classification trees. Subsequently, the GATree system (Papagelis &

[1] Affiliation: Hellenic Open University and Open University of Cyprus, dkalles@acm.org

[2] Affiliation: Department of Computer Engineering and Informatics, University of Patras, Greece, papagel@ceid.upatras.gr

Kalles, 2001) employed a fitness function to trade off tree size (as expressed by the number of leaves) against accuracy in its quest for a good tree.

Genetic algorithms have made an inroad into decision tree induction because they provide an attractive alternative for searching the space of decision trees that can serve as hypothesis for a given data set. While conventional learners, such as CART (Breiman *et al.*, 1984) and C4.5 (Quinlan, 1993) are undoubtedly robust and widely used, it is also a fact that the top-down decisions made early on during decision tree building, such as which attribute to use at the root of the tree, are almost impossible to reverse during subsequent operations. Allowing more alternatives to participate into the decision tree building procedure entails either significant engineering of existing approaches (Esmeir & Markovitch, 2007) or the explicit teaming of classifiers into ensembles to generate more robust results (Breiman, 2001). Genetic algorithms offer a compact and elegant alternative to those approaches, with the added benefit that by manipulating a fitness function, one can inject preference measures that would be otherwise difficult to use and still obtain comparably good accuracy results, with usually significantly smaller trees and an easy to use time-vs-performance trade-off.

Fitness inheritance in genetic algorithms was first explicitly dealt with in the paper by Handley (1994) who explicitly built a relationship graph to represent the associations between members of a population; his approach was thereafter also referred to as fitness case-indexed caching (Ehrenburg, 1996). An alternative notion of inheritance, however, suggested that some individuals need not have their fitness calculated but an approximation should be used instead (Smith *et al.*, 1995). Based on this observation which was backed by limited but promising experiments, several researchers pursued the theoretical justification of the inheritance idea (Sastry *et al.*, 2001), as well as an investigation of techniques that can be used to better approximate the fitness function (Jin, 2005).

This paper furthers the application of genetic algorithms into the problem of building good decision trees by trying to estimate which parts of the fitness function used in the evolution of decision trees can be re-used as efficiently as possible, so that more candidate trees can be examined in the same amount of time. Our goal is to analyze whether such efficiencies are possible in a lossless fashion, namely that the fitness function will be fully calculated based on past data *without* having to revert to approximations.

The rest of the paper is structured in four sections. We first review the basic concepts of genetic evolution of decision trees, also anchoring these concepts to mainstream research strands. Section 3 then shows how the fitness function can be reconstructed in a lossless fashion by re-using parts of decision trees in the current population (an example is included) and then presents an estimate of the expected improvement. Following that, in Section 4 we discuss how the tightening or loosening of some working assumptions might influence our estimates and offer a short discussion on the wider applicability of the lossless fitness inheritance concept. In Section 5 we offer some experimental results to support our arguments and, finally, in Section 6 we conclude and prioritize the directions for further work.

## 2. Genetic evolution of decision trees – a brief review of key issues

We first review the GATree system (Papagelis & Kalles, 2001), which served as the basis for this research, since it produced significantly smaller trees with comparable accuracy, when compared to standard decision tree inducers (Papagelis & Kalles, 2001; Kalles & Pierrakeas, 2006).

GATree first builds a population of decision trees consisting of one node and two leaves. Every decision node has a random chosen attribute-value test. This is done in two steps. First we choose a random attribute. Then, if that attribute is nominal we randomly choose one of its possible values; if it is continuous we randomly pick an integer value belonging to its [*min..max*] range. For leaves, we just pick a random class from the ones available. Mutations operate on single trees (actually, on nodes of single trees) whereas crossovers operate on pairs of trees, after having chosen a crossover node for each of these trees (see Section 3 for examples of these operators at work).

A natural way to assign utility to a decision tree is by using it to classify the training instance-set. Each tree is granted a payoff that is balanced between accuracy and size, allowing for an extra parameter, *x*, to tune their relative weight:

$$payoff(tree_i) = accuracy_i^2 * \frac{x}{size_i^2 + x} \qquad \text{Eq. 1}$$

Note that as an attribute-value test can be used more than one time along a decision path, our higher payoff for smaller trees also "punishes" such replications that result in larger trees (semantically similar to smaller ones). The problem we deal with in this paper is the effective and efficient reuse of calculations for the *accuracy* and *size* quantities.

GATree emphasizes the employment of a *weak* procedural bias that allows the concept learner to consider a (relatively speaking) large number of hypotheses in an efficient manner. By employing global metrics of tree quality, such as size and accuracy, it shifts the focus from "how to induce a tree" (standard, impurity-based induction) to "what criteria an induced tree must satisfy". By setting a policy direction, as opposed to how a policy should be implemented, we achieve a *de facto* decrease in search bias. This reflects the employment of genetic algorithms as an alternative search space navigation mechanism, compared to the conventional top down induction algorithms for decision trees.

Conventional decision tree pruning algorithms, such as error-complexity pruning (Breiman *et al.*, 1984), take into account the size-vs.-accuracy trade-off explicitly by first homing on a decision tree and then working bottom-up towards a pruned version. The same trade-off is accommodated in a top-down fashion by the minimum description length principle (Quinlan & Rivest, 1989), which evaluates each node based on the representation cost of a decision tree *and* the training set errors that this tree incurs. GATree is different to the above generic classes of techniques by holistically evaluating a decision tree as opposed to explicit local searching for improvements. Holistic evaluation is also a central theme in the shift from exhaustive search during learning to anytime learning (Esmeir & Markovitch, 2007). However, genetic evolution has been used in decision tree induction, mainly as a tool in feature selection (Chai *et al.*, 1996; Krętowski & Grześ, 2005; Rokach, 2008) and feature construction (Cantú-Paz & Kamath, 2003; Krętowski, 2004), with some approaches explicitly factoring a size-vs.-accuracy trade-off into the fitness function either by computing it (Krętowski & Grześ, 2005) or by estimating it (Rokach, 2008) using the Vapnik—Chervonenkis dimension on oblivious decision trees; the latter, though less powerful, can guarantee the overall minimisation of the subset of the selected features and are thus easier amenable to theoretical analysis, as they are generated according to a pre-specified order of features.

Of course, using previous calculations as a proxy for the final calculation is a classic *caching* concept so it is not surprising that it emerges in the population dynamics of genetic algorithms to deal with the massive fitness function calculations incurred by some applications.

In genetic algorithms the general problem of caching intermediate results for later use has been mostly treated as an instance of fitness inheritance (Roberts, 2003). In decision trees, the fundamental problem of obtaining a good estimate by not having to process all data has been primarily addressed by processing a subset of the dataset with the view of using the metrics derived for this subset as surrogates for the whole dataset (Catlett, 1992; Musick *et al.*, 1993). A similar approach was to explicitly implement caches that store intermediate splitting heuristic results in the hope that such results can be reused at other parts of a decision tree (Kalles, 1994). This approach has been also independently employed in genetic algorithms by using fitness surrogates to explicitly search for fitness evaluation calculations that can be trimmed (Teller & Andre, 1997; Zhang & Joung, 1999). While the latter approach is close to the concept of fitness inheritance, we believe that the use of surrogates and approximations (both in decision trees and genetic algorithms) has distracted researchers from the more fundamental (and, quite easier, as it turned out) problem of direct optimization that is made possible by the structural (namely, divide-and-conquer) nature of some evolutionary computation problems.

A seminal contribution to such structural population dynamics explicitly built a relationship graph to represent the associations between members of a population (Handley, 1994). Establishing a graph structure for identifying reusable parts of representations has been also central to the concept of Cartesian Genetic Programming (Woodward, 2006) and has also apparently influenced the work on identifying which evolutionary computation calculations are eventually not required (and may be trimmed) if one fixes the expected number of generations for a genetic algorithm experiment (Poli & Langdon, 2006). Note that the latter work also draws on the work by Teller and Andre (1997), referred to in the previous paragraph, thus bridging from another viewpoint genetic algorithms with sub-sampling techniques.

A very interesting approach to caching has been also applied by Rokach (2008) who maintains a cache of previously generated oblivious decision trees associated with the corresponding feature set and, upon observing a particular feature subset that has been genetically selected for feature partitioning, retrieves the corresponding decision tree instead of generating it anew. Rokach's approach is distinct to ours but also quite related at a high level; one could possibly identify traces of duality since Rokach caches decision tree parts and calculates the fitness while we cache fitness parts but calculate the decision trees.

**3. Optimizing genetic operators on decision trees**

We will now show that simple data structures, which can be expected to be used for manipulating decision trees, do suffice for implementing lossless fitness inheritance.

A tree that is the output of a mutation can have its fitness function computed directly from its (one) predecessor's input. Note that there are a limited number of fundamentally different outcomes when a tree is subject to a mutation operator; a description of those appears in Table 1. Still, one might define a new operator by composition (for example, change the class assignment at a leaf *and* the attribute-value test at the root).

Table 1. Fundamental mutation operators

| Operator | Action |
|---|---|
| *Node-Change* | An internal node may have its attribute-value test changed |
| *Leaf-Change* | A leaf node has its class assignment changed |
| *Node-Prune* | An internal node may be marked as a leaf (discarding all structure below) |

We start by observing that one can calculate the accuracy of a given decision tree recursively. During testing, when an instance arrives at a leaf, we can increment a counter of the correctly classified instances at that leaf. Working bottom-up we can then calculate the accuracy for the whole decision tree. The same applies for the calculation of the tree size. The divide-and-conquer formulation of the calculation of these quantities will be central to our optimization.

To see how the mutated tree can have its fitness function computed without having to use the whole data set, let us examine *Node-Prune*. Denote by $I(N)$ the set of instances that have passed through node $N$ (for the sake of simplicity, we will use $N$ to denote both a node and the tree rooted at that node). When the *Node-Prune* operator is applied to node $N$, we know that $I(N)$ does not have to be re-evaluated since the path from the root to $N$ has not changed. However, the accuracy of $N$ has to be re-evaluated according to the new class assignment of (the node-converted-to) leaf $N$. After the re-evaluation is complete, the accuracy of the whole tree can be recomputed by only updating accuracy counters along the path to the root. In essence, a mutation indicates a node that is *stale* and needs reexamination.

It is easy to see that for the other two operators, the same technique applies: a mutation only causes the mutated node to have its accuracy and size component re-calculated. If the node is internal, that means having to recalculate accuracies and sizes recursively for that portion of the tree that lies below the mutated node.

We now turn to show how a tree that is the output of a crossover can have its fitness function computed directly from its (two) predecessors' input. We first note that there is just one crossover operator (see Figure 1 for an example).

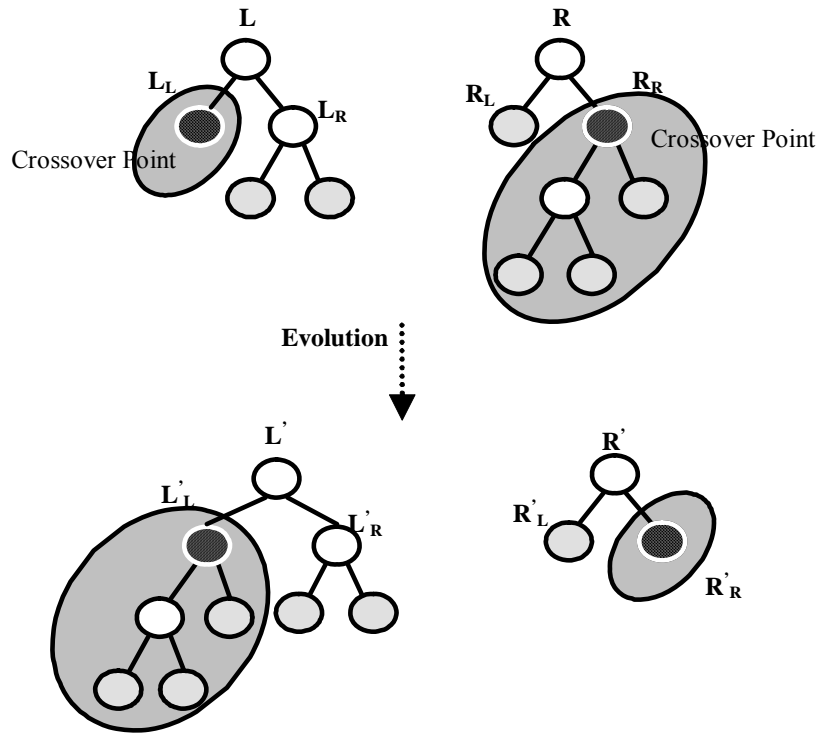

Figure 1: An example of the crossover operator.

Since the paths from the roots to the corresponding nodes have not changed, it is the case again that $I(L_L)=I(L'_L)$ and that $I(R_R)=I(R'_R)$. If one of the crossover points is a tree root, we simply note that the new tree triggers an examination of the whole data set to calculate the accuracy and the previous observation about stale nodes apply. A special case is when crossover occurs at *both* tree roots; in that case a simple swapping of fitness indicators suffices.

The above show that the observation of which nodes are stale and warrant re-examination is again the key to treating crossovers.

**3.1. A brief example to demonstrate improvement**

We now present a brief example to clarify the points above. Consider a sample data set with four instances, three binary attributes and one binary class, as shown in Table 2.

Table 2. A sample data set

|       | $A_1$ | $A_2$ | $A_3$ | Class |
|-------|-------|-------|-------|-------|
| $I_1$ | N     | N     | Y     | Y     |
| $I_2$ | N     | Y     | N     | N     |
| $I_3$ | Y     | N     | N     | N     |
| $I_4$ | Y     | Y     | Y     | Y     |

Assume that at some point of the evolution we are considering the decision trees shown in Figure 2. All left branches correspond to a value of *N* for the tested attribute. Instances shown below leaves indicate the classification path for the corresponding instance (from the tree root to that leaf). Strikethrough instances indicate a classification error. Arrows show the nodes that have been selected for crossover.

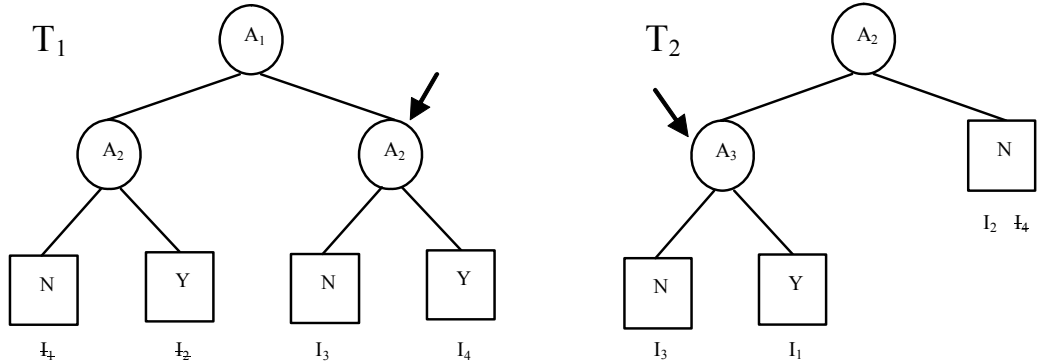

Figure 2: A snapshot of the evolution: trees $T_1$ and $T_2$.

By setting the *x* factor of the fitness formula to 1, we obtain:

$$F_1 = (0 + \underline{2})^2 * 1 / ((2 + \underline{2})^2 + 1) = 4 / 17 \qquad F_2 = (\underline{2} + 1)^2 * 1 / ((\underline{2} + 1)^2 + 1) = 9 / 10$$

In the above derivation we underline the quantities that will have to be recomputed in the next evolution step. For example, the (0 + 2) figure in the derivation of $F_1$ indicates that only 2 instances are correctly classified and these are due to the right branch of the root and that in the next evolution step the left sub-tree will not have to have its accuracy recomputed. Tree size is measured by the number of leaves.

After the crossover, the resulting trees are shown in Figure 3 (redundant nodes are not collapsed).

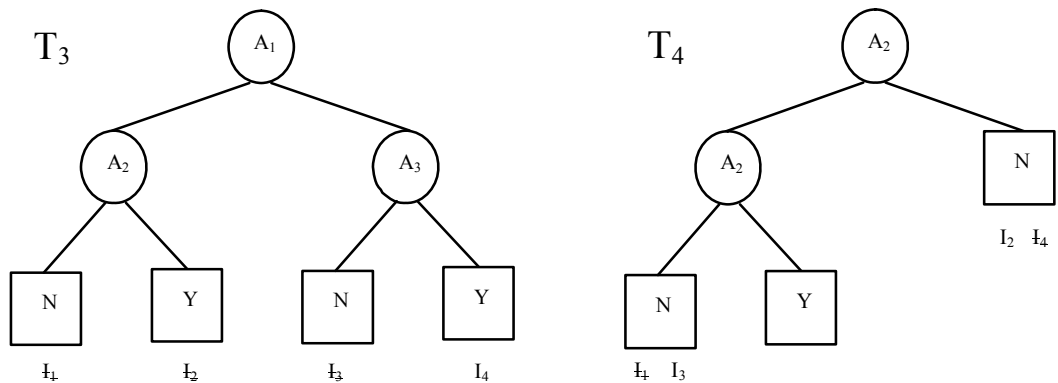

Figure 3: A new snapshot of the evolution: trees $T_3$ and $T_4$.

For the new fitness values we observe that $F_4$ differs from $F_2$ only because the left branch of $T_4$ now creates an extra error in the classification of $I_1$:

$$F_3 = 2^2 * 1 / (4^2 + 1) = 4 / 17 \qquad F_4 = (\underline{1} + 1)^2 * 1 / ((\underline{2} + 1)^2 + 1) = 4 / 10$$

The cautious reader might observe at this point that the extra cost of recursively adding accuracy counters to calculate the tree accuracy might contribute to a non-negligible delay in the evolution and that this delay should be factored in any attempt to estimate speed-up. We point out that the fundamental algorithm already incurs this cost by having to recursively compute the size of the tree. Since the fitness function takes size into account, the traversal of the tree to calculate its size is an operation that gets done anyway.

Piggy-backing the accuracy calculation in this traversal all but eliminates the overhead, since the only extra data we need is the set of instances at each node.

**3.2. Expected improvement**

Since dealing with generic binary decision trees is too complicated, we examine two types of trees: the complete tree and the linear tree (see Figure 4 for an example). A complete tree contains all internal nodes and leaves up to a given depth, whereas a linear tree only contains one internal node and one leaf per level. Complete binary decision trees are quite unlikely to occur in a practical machine learning context (linear ones are considerably more likely, also due to their intuitive layout). However, complete trees offer theoretical insight and have been also studied as an example of the very difficult XOR concept (Pagallo, 1989). We refer to both types of trees as bounding cases (extreme trees) in the sense that all possible trees with the same number of leaves are no deeper than the linear tree and no shallower than the complete tree.

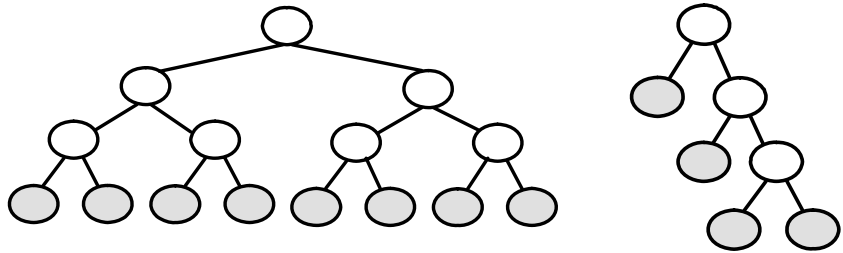

Figure 4: Examples of extreme binary decision trees: a complete (left) and a linear one (right). Darker nodes indicate leaves whereas blank nodes are internal nodes.

There is a simple argument to appreciate why the above constraints on the type of decision trees we study do not in fact dilute the potential of the approach. That is to view such trees as the output of a genetic operator and retrospectively ask where, on any such given tree, a node might have been rendered stale. In fact, if we had explicitly considered such a tree as input only (and not as output), a legitimate question would be whether we should also discuss the probability that the tree *type* would not change after an operator had acted (for example, that a complete tree would remain complete). With better analytical skills, one might be able to compute such probability even in generic trees (Knessl & Szpankowski, 2006); but that is beyond the scope of this paper.

**3.2.1. Analyzing complete trees**

On notation issues, we start by noting that a complete binary decision tree has $k+1$ levels, in the range $0..k$. The $0^{th}$ level is the tree root. The $k^{th}$ level consists of leaves only; furthermore, no leaves are to be found elsewhere in the tree. Let us also assume that the tree has $n$ instances, evenly distributed among the $2^k$ leaves. Thus, the tree contains $2^{k+1}+1$ nodes (including internal nodes and leaves).

We now define the *node-instance-check* as the unit of operation that processes an instance at a node. Such an operation at an internal node is the equivalent of a test with the attribute-value at that node. An operation at a leaf is the examination of the class at that leaf. Hence, $k+1$ is the (maximum) number of node-instance-check (hereafter, simply referred to as operations) required by any instance as it is sent down the tree.

Now, assume we have a node at level $i$. Such a node covers $n/2^i$ instances (via all nodes below it), all the way down to the leaves. These account for $(n/2^i)(k+1)$ operations (since each instance has to start being processed at the root, anyway).

We are now ready to frame the problem: if a node at level $i$ is marked as stale, what is the estimated cost of updating its data structures, so that size and accuracy are correctly calculated?

We start by calculating $Cost_0$, the cost of classifying all instances through the whole decision tree. Since each of $n$ instances consumes $k+1$ operations, $Cost_0 = n(k+1)$.

Updating at level $i$ incurs the cost of re-classifying that part of the data set that passes through the stale node, $Cost_i = (n/2^i)(k+1)$.

The expected new cost is $Cost_i / Cost_0 = 1/2^i$ (note that the number of instances, $n$, has been factored out).

The cost ratio (denoted henceforth by $R$) has to be averaged over all nodes of the tree. This factors in the probability that a node at level $i$ will be selected, which is $2^i/(2^{k+1}-1)$. Note that we average uniformly across all nodes reflecting that each one has the same probability to be selected for crossover.

$$R_{complete} = \sum_{i=0}^{k}\left(\frac{2^i}{2^{k+1}-1}\right)\left(\frac{1}{2^i}\right) = \frac{k+1}{2^{k+1}-1} \qquad \text{Eq. 2}$$

For a depth-3 complete tree with 8 leaves, we get $R_{complete,k=3}<0.27$. Similarly, $R_{complete,k=9}<0.01$. Note that improvement is near-exponential and quite substantial even for small trees too.

**3.2.2. Analyzing linear trees**

We now turn to linear binary decision trees. These are skewed in the sense that, at each level, there is one leaf. The last level contains two leaves only. We study linear trees because they offer insight into the pessimistic cases.

Again, a linear binary decision tree has $k+1$ levels, in the range $0..k$. The $0^{th}$ level is the tree root. Let us also assume that the tree has $n$ instances, evenly distributed among the $k+1$ leaves, accounting for $n/(k+1)$ instances per leaf. Thus, the tree contains $2k+1$ nodes (including internal nodes and leaves). Yet again, $k+1$ is the (maximum) number of node-instance-check operations required by any instance as it is sent down the tree.

Now, assume we have a node at level $i$. Such a node covers $n - i\left(\frac{n}{k+1}\right) = n\left(\frac{k-i+1}{k+1}\right)$ instances.

However, for this type of tree, the number of operations consumed by an instance depends on which leaf it will be found (as opposed to the complete tree, where all instances travel the same path length).

Note that the lowest-level leaves will require $k+1$ operations for each of their instances. Going one level up, a leaf will require $k$ operations for each of its instances. In general, a leaf at level $i$ will require $i+1$ operations (this also holds for $i=1$; an instance at the topmost leaf consumes 2 operations).

We will now calculate $Cost_0$; therein we factor the cost of the lowest-rightmost leaf and then we sum the costs of the other leaves as we move towards the root.

$$Cost_0 = \frac{n}{k+1}(k+1) + \frac{n}{k+1}\sum_{i=1}^{k}(i+1) = n + \frac{n}{k+1}\left(k + \sum_{i=1}^{k}i\right) = n\left(1 + \frac{k}{k+1} + \frac{k}{2}\right) \qquad \text{Eq. 3}$$

To calculate $Cost_i$ we must differentiate between an internal node and a leaf.

The leaf case is straightforward:

$$Cost_{i,leaf} = n\frac{i+1}{k+1} \qquad \text{Eq. 4}$$

For the internal node case, we observe that the cost is simply the formula for $Cost_0$ (see Eq. 3) applied on a tree of $k-i$ levels plus the cost of having all instances covered by the modified node actually reach that node first (each incurring the same overhead up to that node, of course).

$$Cost_{i,node} = n\left(1 + \frac{k-i}{k-i+1} + \frac{k-i}{2}\right) + ni\frac{i+1}{k+1} \qquad \text{Eq. 5}$$

The cost ratio has to be averaged over all nodes of the tree. Every node has a probability of $1/(2k+1)$ to be selected. Since there is one node at level 0 and two nodes for each level thereafter, we get:

$$R_{linear} = \frac{1}{2k+1}\frac{Cost_0}{Cost_0} + \frac{1}{2k+1}\sum_{i=1}^{k}\frac{Cost_{i,leaf}}{Cost_0} + \frac{1}{2k+1}\sum_{i=1}^{k}\frac{Cost_{i,node}}{Cost_0} \qquad \text{Eq. 6}$$

We now note that $Cost_i$ is a monotonously increasing function of $i$, so we approximate its calculation[3]:

$$R_{linear} \le R_{est} =$$
$$= \frac{1}{2k+1} + \frac{1}{2k+1}\left(\frac{Cost_{i=1,leaf} + Cost_{i=k,leaf}}{Cost_0}\frac{1}{2}k\right) + \frac{1}{2k+1}\left(\frac{Cost_{i=1,node} + Cost_{i=k,node}}{Cost_0}\frac{1}{2}k\right) \qquad \text{Eq. 7}$$

After some algebraic manipulations we obtain:

$$R_{est} = \frac{1}{2k+1} + \frac{1}{2k+1}\frac{k^2+3k}{k^2+5k+2} + \frac{1}{2k+1}\left(\frac{3k^3+8k^2-k-2}{2(k^2+5k+2)}\right) \le$$
$$\le \frac{2}{2k+1} + \frac{1}{2k+1}3\frac{k-1}{2} = \frac{3k+1}{4k+2} \qquad \text{Eq. 8}$$

Now, comparing Eq. 2 (complete binary decision tree) to Eq. 8, we observe that improvement is not as impressive, but it still converges to about 25% ($1 - \frac{3}{4}$).[4] However, as opposed to the complete tree case, it is now the deep trees that demonstrate smaller improvement, while shallow trees do much better.

## 4. On the validity and the implications of the results

We now first reflect on the accuracy of the estimation and then proceed to discuss the wider applicability of our observation in other fitness inheritance settings.

We stress again that the extension to the tree data structure is minimal. It only consists of an array of instance indices per node. Quite importantly, when a stale node is processed, no nodes on the way up to the root need have their data structures changed (hence, incurring possibly expensive array union operations) since the instance sets at those nodes are guaranteed not to have changed.

However, space may be expensive. The amount of space taken up by the instance indices in each node can be shown to be $O(kn)$ for the two variants that we studied. If we take into account that these indices have to be stored in each tree, the resulting cost may grow at the expense of our ability to deal with huge datasets – we remind the reader that the basic algorithm does not store any dataset related information. We can improve substantially upon that $O(kn)$ cost by employing the argument on the piggybacking of operations onto the tree size calculation (as presented just before Section 3.2). If we store the instance indices *only* in the leaves, we can recursively build our instance index at any node by concatenating the instance indices from below. Simple concatenation suffices since we have no duplicates and since no ordering information is required. This trims the storage cost to $O(n)$.

---

[3] Note that $Cost_{i,leaf}$ is a linear function, hence the approximation refers only to $Cost_{i,node}$. We have calculated the error of the approximation at about 8% for $k=8$ (but the error increases with $k$). However, the approximation is pessimistic and we refer the reader to section 4 for a more detailed discussion on whether such pessimism is really warranted, offering a further at least double-fold improvement.

[4] We again refer the reader to section 4 where this estimate is substantially improved by partially avoiding some reclassification effort.

The cautious reader might observe that when a decision tree substitutes part of itself with part of another decision tree, there is no guarantee that the new decision tree will be of the same layout. However, there is a strong reason why one might not need to venture into factoring this observation explicitly in the analysis (we also believe that the complicated mathematics is unlikely to provide any further insight). The reason is that, on average, the probability that the new part will result in a larger tree is balanced by the probability that the new part will be smaller and hence result in a smaller tree. As a matter of fact, since the two parts will be simply exchanged and since both tree roots refer to the same instances, the extra operations that may be incurred at one side will be saved at the other side.

Furthermore, it is instructive to see why the derivations of the potential savings, as detailed in Section 3.2, are pessimistic. Still, one might decide to plan with these pessimistic estimates, to safely offset the overhead cost of housekeeping the data structures.

Recall that for the complete binary decision tree we assumed that updating at level $i$ incurs the cost of re-classifying that part of the data set that passes through the stale node, $Cost_i = (n/2^i)(k+1)$. This is pessimistic since we *only* need to reclassify at the node at level $i$ (and below), thus incurring a cost of $Cost'_i = (n/2^i)(k-i+1)$. Factoring the modified cost into our calculations would result in a further speed-up by a factor of 2.

For the linear binary decision tree the cost would be:

$$Cost'_{i,node} = n\left(1 + \frac{k-i}{k-i+1} + \frac{k-i}{2}\right) \qquad \text{Eq. 9}$$

The associated savings ratio is $R'_{est} \leq \frac{k+5}{4k+2}$, which converges to 0.25 accounting for a 3-fold increase in speed-up over the pessimistically calculated quantity.

Figure 5 demonstrates a graphical plot of the two $1-R'$ quantities, showing the fraction of saved node-instance-checks as a function of tree depth. To enhance the quality of the plot the *x* axis is shown in logarithmic scale. Note that even the shallowest trees demonstrate significant improvement and that the maximum attainable improvement is early reached.

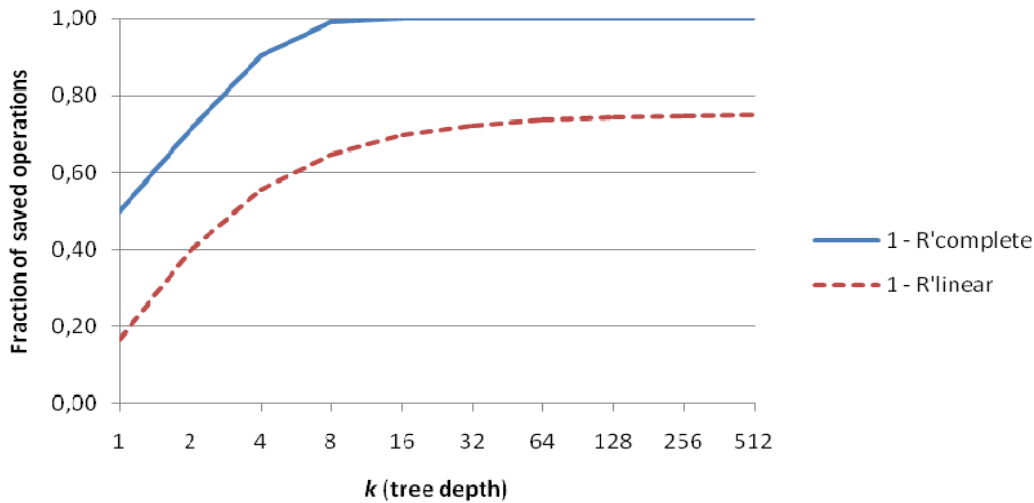

Figure 5: A graphical plot of saved node-instance-checks as a function of tree depth.

One could argue that the definition of the linear tree should allow leaves which are nearer the tree root to accommodate more instances than the leaves that are near the maximum tree depth. We briefly analyze this alternative here.

Again, such a *weighted* linear binary decision tree has $k+1$ levels, in the range $0..k$, with the tree root at the $0^{th}$ level. Now, a leaf at level $i$ contains $n/(2^i)$ instances out of $n$ instances in total.

We first calculate $Cost_0$ by explicitly accounting for one of the two lowest leaves and then summing the costs of the other leaves as we move towards the root.

$$Cost_0 = \frac{n}{2^k}(k+1) + n\sum_{i=1}^{k}\frac{i+1}{2^i} \leq n\left(\frac{k+1}{2^k} + k + 2\right) \qquad \text{Eq. 10}$$

We acknowledge that $k+2$ in the derivation in Eq. 10 is a weak upper bound, but we also note that this does not substantially influence our final estimates.

For the rest of the nodes, we derive:

$$Cost_{i,leaf} = n\frac{k-i+1}{2^i} \qquad \text{Eq. 11}$$

$$Cost_{i,node} \leq n\left(\frac{k-i+1}{2^{k-i}} + k - i + 2\right) \qquad \text{Eq. 12}$$

We now derive approximations for $\dfrac{Cost_{i,leaf}}{Cost_0}$ and $\dfrac{Cost_{i,node}}{Cost_0}$ to facilitate the math:

$$\frac{Cost_{i,leaf}}{Cost_0} \leq \frac{n\dfrac{k-i+1}{2^i}}{n\left(\dfrac{k+1}{2^k}+k+2\right)} = \frac{2^k(k-i+1)}{2^i(k+1+2^k(k+2))} \leq \frac{2^k(k-i+1)}{2^i 2^k(k+2)} = \frac{k-i+1}{2^i(k+2)} \qquad \text{Eq. 13}$$

$$\frac{Cost_{i,node}}{Cost_0} \approx \frac{n\left(\dfrac{k-i+1}{2^{k-i}}+k-i+2\right)}{n\left(\dfrac{k+1}{2^k}+k+2\right)} = \frac{2^k(k-i+1+2^{k-i}(k-i+2))}{2^{k-i}(k+1+2^k(k+2))} \leq$$

$$\leq \frac{(k-i+2)(2^k+2^{2k-i})}{(k+1)(2^{k-i}+2^{2k-i})} \leq \frac{2^{2k-i+1}(k-i+2)}{2^{2k-i}(k+1)} = \frac{2(k-i+2)}{(k+1)} \qquad \text{Eq. 14}$$

The overall ratio has to be averaged over all nodes of the tree:

$$R_{weighted \land linear} = \frac{1}{2k+1}\frac{Cost_0}{Cost_0} + \frac{1}{2k+1}\sum_{i=1}^{k}\frac{Cost_{i,leaf}}{Cost_0} + \frac{1}{2k+1}\sum_{i=1}^{k}\frac{Cost_{i,node}}{Cost_0} \leq$$

$$\leq \frac{1}{2k+1} + \frac{1}{2k+1}\sum_{i=1}^{k}\frac{k-i+1}{2^i(k+2)} + \frac{1}{2k+1}\sum_{i=1}^{k}\frac{2(k-i+2)}{(k+1)} =$$

$$= \frac{1}{2k+1} + \frac{1}{2k+1}\frac{1}{k+2}\left(\sum_{i=1}^{k}\frac{k}{2^i} + \sum_{i=1}^{k}\frac{1}{2^i} - \sum_{i=1}^{k}\frac{i}{2^i}\right) + \frac{2}{2k+1}\frac{1}{k+1}\left(\sum_{i=1}^{k}(k+2) - \sum_{i=1}^{k}i\right) \leq \quad \text{Eq. 15}$$

$$\leq \frac{1}{2k+1} + \frac{1}{2k+1}\frac{1}{k+2}k + \frac{1}{2k+1}\frac{1}{k+1}k(k+3) =$$

$$= \frac{1}{2k+1}\left(1 + \frac{k}{k+2} + \frac{k(k+3)}{k+1}\right) \leq \frac{1}{2k+1}(1+1+(k+3)) = \frac{k+5}{2k+1}$$

Note that the above approximation is grossly pessimistic for small values of *k* but for those values actual improvement would be straightforward to calculate. The asymptotic improvement is 50%.

**4.1. A brief speculation on further applicability**

The formulation of vital tree statistics (size, accuracy) in terms of recursive calculations in a divide-and-conquer fashion suggest that it might be interesting to investigate whether such speed-up results may be found in other problems where the evaluation of the fitness function can be cast as a divide-and-conquer problem.

To gain some insight into this potential we briefly review a case into the other extreme. Suppose that we were evolving conventional feed-forward neural networks instead of decision trees. Since each change in a neural network would affect *at least one* weight, *every* instance would be liable to have its whole NN-based classification changed. Neural networks are not designed for divide-and-conquer computations, hence there is minimal potential in using past fitness calculations for the lossless reconstruction of fitness in future populations. At this point we should remind the reader that it is the lossless reconstruction we are interested in; approximation techniques have been effectively used in domains where this assumption is relaxed (Pelikan & Sastry, 2004).

The lack of interdependence of local problems in a divide-and-conquer formulation has been also indirectly explored in research on how to optimize the convergence of genetic algorithms. For example, it has been noted that fitness caching can be employed only where evaluations have no side effects (Roberts, 2003). The conclusion has seemed to be that for all but simple cases, the problem is hard (see Jansen (1999) and references therein for an overview). Not surprisingly, using the NK-model taxonomy (see Altenberg (1997) for a review), it is straightforward to see why the calculation of size and accuracy sets the decision trees problem that we studied at the easiest end of the scale (*n*0), since no interactions occur between the genome parts. Independently, fitness functions that can be piecewise computed have been recast as Walsh polynomials for faster convergence (Linton, 2004). However, it is the piecewise computation that seems to be the underlying assumption in that line of research; we expect that divide-and-conquer fitness functions to be cast in that representation without any problems.

**5. An indicative experimental evaluation**

We modified GATree to incorporate the above changes and we tested it on 12 data sets from the Machine Learning Repository (Asuncion & Newman, 2007).

For each data set we ran 8 experiments, each with an increasing number of generations and populations per run (so, column "700" indicates experiments with 700 generations and 700 trees per generation). For each data set and each experiment therein we report how long it took to build the tree in the "old" approach, how long it took to build the tree in the proposed "new" approach, and what part of instances did *not* have to be re-evaluated (re-classified) due to the implementation of the "new" approach.

To increase the credibility of the results we repeated the above procedure 4 times, with different combinations of the mutation rate and the *x*-factor (see Section 2)[5]:

- For a mutation rate of 0.5 (mutation rate is the mutation probability of a standard genetic algorithm approach)
    o the *x*-factor was set at $10^4$ (see Table 4),
    o the *x*-factor was set at $10^4$ at the start of the evolution and increased linearly up to $10^5$ at the end of the evolution (see Table 5).
- For a mutation rate of 0.01
    o the *x*-factor was set at $10^4$ (see Table 6),
    o the *x*-factor was set at $10^4$ at the start of the evolution and increased linearly up to $10^5$ at the end of the evolution (see Table 7).

An average of the savings per data set and per configuration is shown in Table 3 (note that all savings are is reported as a percentage number – the % symbol is omitted).

Table 3. An average of the savings across experiments

| Dataset | x = 10000 | | x in [10000..100000] | |
|---|---|---|---|---|
| | m = 0.5 | m = 0.01 | m = 0.5 | m = 0.01 |
| balance-scale | 56,25 | 66,64 | 69,01 | 73,32 |
| zoo | 56,73 | 66,18 | 66,36 | 67,75 |
| credit-a | 35,16 | 45,77 | 53,55 | 59,32 |
| lymph | 56,13 | 63,57 | 69,74 | 69,48 |
| glass | 50,91 | 59,65 | 63,56 | 64,88 |
| soybean | 46,81 | 60,45 | 47,37 | 60,05 |
| vote | 29,42 | 42,78 | 37,48 | 47,28 |
| anneal | 37,13 | 44,66 | 52,45 | 45,15 |
| crime | 39,70 | 45,73 | 55,73 | 51,87 |
| kr-vs-kp | 37,16 | 42,15 | 45,69 | 45,67 |
| breast | 55,40 | 64,51 | 68,68 | 70,65 |
| multiplexor | 54,58 | 59,78 | 71,99 | 69,90 |

---

[5] All other GATree parameters were set at their default values, but each experiment was seeded at a different random number seed. We opted to test with different configurations as opposed to running the same experiment several times, each with a new random seed, to indicate the robustness potential of the proposed approach.

Table 4. A savings-only comparison with mutation rate at 0.5 and $x$ at $10^4$

| Dataset | | 100 | 200 | 300 | 400 | 500 | 600 | 700 | 800 |
|---|---|---|---|---|---|---|---|---|---|
| balance-scale | Old (sec) | 0,98 | 4,34 | 9,84 | 17,97 | 27,69 | 34,90 | 59,71 | 72,91 |
| | New (sec) | 0,71 | 2,47 | 5,97 | 10,79 | 18,20 | 25,06 | 35,53 | 44,55 |
| | Train Savings | 49,54 | 56,91 | 58,63 | 54,15 | 59,35 | 55,50 | 59,70 | 56,26 |
| zoo | Old (sec) | 0,52 | 2,33 | 5,76 | 9,50 | 16,06 | 23,88 | 32,36 | 45,58 |
| | New (sec) | 0,39 | 1,80 | 4,68 | 7,63 | 13,48 | 16,90 | 24,94 | 35,60 |
| | Train Savings | 55,48 | 60,00 | 54,99 | 54,65 | 57,91 | 52,39 | 57,20 | 61,20 |
| credit-a | Old (sec) | 1,66 | 6,81 | 15,61 | 27,27 | 43,58 | 61,80 | 86,98 | 111,24 |
| | New (sec) | 0,86 | 4,18 | 8,36 | 17,98 | 27,17 | 33,99 | 49,84 | 66,33 |
| | Train Savings | 29,08 | 35,69 | 40,56 | 33,30 | 32,47 | 43,31 | 31,62 | 35,24 |
| lymph | Old (sec) | 0,60 | 2,90 | 6,20 | 13,04 | 23,50 | 29,87 | 44,97 | 56,55 |
| | New (sec) | 0,38 | 1,82 | 4,16 | 8,43 | 13,38 | 21,68 | 25,17 | 38,83 |
| | Train Savings | 51,18 | 56,30 | 46,45 | 60,54 | 52,95 | 61,54 | 56,14 | 63,94 |
| glass | Old (sec) | 1,05 | 4,73 | 12,62 | 20,30 | 36,39 | 59,23 | 84,46 | 112,83 |
| | New (sec) | 0,63 | 2,38 | 6,65 | 12,00 | 18,89 | 30,20 | 31,23 | 58,10 |
| | Train Savings | 42,86 | 50,16 | 49,83 | 52,02 | 54,35 | 48,68 | 60,62 | 48,79 |
| soybean | Old (sec) | 0,38 | 1,71 | 3,12 | 5,17 | 8,54 | 12,84 | 17,91 | 22,29 |
| | New (sec) | 0,25 | 1,51 | 2,53 | 4,59 | 7,48 | 11,59 | 14,72 | 19,01 |
| | Train Savings | 46,98 | 56,04 | 45,00 | 46,84 | 45,44 | 46,70 | 43,65 | 43,85 |
| vote | Old (sec) | 0,81 | 3,30 | 7,24 | 12,83 | 21,33 | 28,66 | 41,91 | 54,05 |
| | New (sec) | 0,60 | 2,47 | 5,47 | 9,67 | 15,72 | 21,65 | 30,01 | 40,15 |
| | Train Savings | 27,88 | 29,93 | 29,59 | 29,55 | 30,11 | 29,30 | 29,64 | 29,37 |
| anneal | Old (sec) | 5,89 | 18,45 | 55,60 | 98,84 | 155,75 | 165,55 | 308,79 | 368,76 |
| | New (sec) | 3,46 | 12,71 | 34,06 | 59,47 | 102,35 | 133,83 | 165,07 | 161,76 |
| | Train Savings | 36,88 | 34,19 | 33,89 | 34,50 | 34,40 | 31,68 | 33,89 | 57,62 |
| crime | Old (sec) | 0,70 | 4,03 | 9,13 | 14,42 | 21,61 | 32,52 | 55,32 | 61,77 |
| | New (sec) | 0,60 | 2,43 | 6,54 | 11,03 | 17,16 | 27,65 | 32,39 | 47,82 |
| | Train Savings | 28,65 | 43,53 | 33,45 | 48,77 | 30,03 | 38,32 | 53,43 | 41,42 |
| kr-vs-kp | Old (sec) | 5,09 | 19,46 | 43,86 | 76,66 | 154,44 | 193,63 | 225,69 | 400,59 |
| | New (sec) | 3,82 | 11,49 | 28,37 | 58,19 | 105,02 | 143,92 | 157,79 | 235,42 |
| | Train Savings | 27,26 | 54,45 | 49,76 | 29,13 | 28,93 | 28,96 | 49,92 | 28,89 |
| breast | Old (sec) | 0,67 | 3,79 | 8,50 | 17,50 | 24,28 | 37,39 | 55,97 | 65,66 |
| | New (sec) | 0,41 | 1,80 | 4,30 | 8,17 | 11,49 | 21,83 | 28,02 | 37,08 |
| | Train Savings | 42,81 | 53,88 | 53,69 | 55,37 | 53,54 | 58,83 | 63,45 | 61,62 |
| multiplexor | Old (sec) | 0,36 | 1,66 | 4,12 | 8,91 | 16,23 | 9,62 | 30,47 | 34,34 |
| | New (sec) | 0,28 | 1,46 | 3,34 | 7,20 | 6,17 | 9,18 | 30,15 | 27,11 |
| | Train Savings | 50,59 | 50,99 | 56,33 | 65,01 | 42,22 | 41,54 | 66,33 | 63,62 |

Table 5. A savings-only comparison with mutation rate at 0.5 and $x$ in $[10^4..10^5]$

| Dataset | | 100 | 200 | 300 | 400 | 500 | 600 | 700 | 800 |
|---|---|---|---|---|---|---|---|---|---|
| balance-scale | Old (sec) | 1,32 | 6,33 | 12,07 | 26,30 | 31,29 | 60,36 | 85,95 | 120,73 |
| | New (sec) | 0,86 | 3,50 | 10,64 | 17,34 | 25,17 | 41,68 | 54,26 | 63,35 |
| | Train Savings | 57,58 | 68,88 | 68,12 | 71,59 | 73,23 | 68,51 | 72,66 | 71,53 |
| zoo | Old (sec) | 0,57 | 3,06 | 7,31 | 12,79 | 21,03 | 32,99 | 30,80 | 59,56 |
| | New (sec) | 0,38 | 2,59 | 7,65 | 14,16 | 16,11 | 23,58 | 38,56 | 41,07 |
| | Train Savings | 55,29 | 68,52 | 71,15 | 73,56 | 62,16 | 65,26 | 68,95 | 65,98 |
| credit-a | Old (sec) | 1,86 | 8,80 | 21,97 | 31,57 | 60,01 | 94,08 | 145,63 | 198,16 |
| | New (sec) | 0,96 | 4,36 | 9,16 | 18,54 | 35,38 | 41,35 | 49,34 | 73,52 |
| | Train Savings | 50,26 | 50,80 | 56,75 | 52,12 | 55,92 | 53,36 | 56,62 | 52,58 |
| lymph | Old (sec) | 0,79 | 4,45 | 10,33 | 20,25 | 31,77 | 43,95 | 66,50 | 95,87 |
| | New (sec) | 0,77 | 3,09 | 7,16 | 15,21 | 26,99 | 39,25 | 46,31 | 60,45 |
| | Train Savings | 65,84 | 65,68 | 67,20 | 72,68 | 72,41 | 73,63 | 71,52 | 68,94 |
| glass | Old (sec) | 0,99 | 7,63 | 20,63 | 42,80 | 58,09 | 135,50 | 109,80 | 157,41 |
| | New (sec) | 0,72 | 4,00 | 8,86 | 20,28 | 54,46 | 85,45 | 72,33 | 102,46 |
| | Train Savings | 55,57 | 59,44 | 69,06 | 65,32 | 60,27 | 62,03 | 67,68 | 69,13 |
| soybean | Old (sec) | 0,35 | 1,70 | 3,71 | 6,21 | 9,48 | 13,36 | 19,11 | 24,07 |
| | New (sec) | 0,30 | 1,63 | 3,18 | 5,06 | 8,42 | 11,48 | 15,94 | 20,63 |
| | Train Savings | 48,67 | 53,17 | 47,96 | 47,82 | 46,71 | 45,18 | 45,30 | 44,17 |
| vote | Old (sec) | 1,14 | 5,35 | 7,48 | 13,20 | 21,58 | 29,38 | 61,14 | 53,06 |
| | New (sec) | 0,60 | 2,46 | 6,92 | 10,27 | 19,59 | 21,90 | 30,52 | 40,95 |
| | Train Savings | 28,87 | 29,90 | 61,04 | 29,76 | 61,66 | 29,32 | 29,82 | 29,44 |
| anneal | Old (sec) | 5,79 | 23,44 | 69,81 | 100,72 | 172,32 | 221,61 | 323,89 | 451,75 |
| | New (sec) | 3,11 | 9,14 | 23,19 | 66,93 | 58,41 | 128,82 | 167,12 | 236,58 |
| | Train Savings | 46,70 | 67,04 | 65,94 | 39,56 | 64,65 | 45,24 | 41,72 | 48,71 |
| crime | Old (sec) | 0,71 | 4,38 | 13,07 | 21,02 | 39,41 | 56,13 | 75,54 | 114,79 |
| | New (sec) | 0,62 | 2,94 | 7,40 | 15,04 | 24,70 | 28,46 | 48,66 | 70,62 |
| | Train Savings | 31,60 | 60,04 | 62,76 | 62,08 | 61,64 | 48,96 | 59,80 | 58,99 |
| kr-vs-kp | Old (sec) | 6,19 | 25,95 | 46,67 | 81,79 | 190,17 | 268,98 | 238,63 | 442,76 |
| | New (sec) | 3,85 | 11,97 | 36,45 | 61,91 | 73,88 | 111,18 | 164,00 | 157,98 |
| | Train Savings | 34,07 | 59,52 | 29,54 | 29,35 | 57,32 | 59,56 | 29,03 | 67,12 |
| breast | Old (sec) | 1,00 | 6,11 | 13,23 | 29,56 | 43,40 | 61,38 | 76,18 | 119,06 |
| | New (sec) | 0,64 | 2,80 | 5,69 | 11,67 | 19,30 | 29,98 | 42,58 | 67,31 |
| | Train Savings | 60,54 | 63,00 | 66,88 | 72,79 | 70,31 | 70,06 | 71,93 | 73,96 |
| multiplexor | Old (sec) | 0,72 | 3,23 | 2,48 | 17,38 | 13,28 | 37,26 | 59,08 | 59,90 |
| | New (sec) | 0,52 | 2,53 | 6,37 | 9,84 | 22,92 | 32,65 | 44,45 | 59,76 |
| | Train Savings | 59,59 | 67,91 | 73,59 | 71,77 | 75,62 | 74,67 | 77,35 | 75,40 |

Table 6. A savings-only comparison with mutation rate at 0.01 and $x$ at $10^4$

| Dataset | | 100 | 200 | 300 | 400 | 500 | 600 | 700 | 800 |
|---|---|---|---|---|---|---|---|---|---|
| balance-scale | Old (sec) | 0,98 | 4,34 | 9,84 | 17,97 | 27,69 | 34,90 | 59,71 | 72,91 |
| | New (sec) | 0,71 | 2,47 | 5,97 | 10,79 | 18,20 | 25,06 | 35,53 | 44,55 |
| | Train Savings | 49,54 | 56,91 | 58,63 | 54,15 | 59,35 | 55,50 | 59,70 | 56,26 |
| zoo | Old (sec) | 0,52 | 2,33 | 5,76 | 9,50 | 16,06 | 23,88 | 32,36 | 45,58 |
| | New (sec) | 0,39 | 1,80 | 4,68 | 7,63 | 13,48 | 16,90 | 24,94 | 35,60 |
| | Train Savings | 55,48 | 60,00 | 54,99 | 54,65 | 57,91 | 52,39 | 57,20 | 61,20 |
| credit-a | Old (sec) | 1,66 | 6,81 | 15,61 | 27,27 | 43,58 | 61,80 | 86,98 | 111,24 |
| | New (sec) | 0,86 | 4,18 | 8,36 | 17,98 | 27,17 | 33,99 | 49,84 | 66,33 |
| | Train Savings | 29,08 | 35,69 | 40,56 | 33,30 | 32,47 | 43,31 | 31,62 | 35,24 |
| lymph | Old (sec) | 0,60 | 2,90 | 6,20 | 13,04 | 23,50 | 29,87 | 44,97 | 56,55 |
| | New (sec) | 0,38 | 1,82 | 4,16 | 8,43 | 13,38 | 21,68 | 25,17 | 38,83 |
| | Train Savings | 51,18 | 56,30 | 46,45 | 60,54 | 52,95 | 61,54 | 56,14 | 63,94 |
| glass | Old (sec) | 1,05 | 4,73 | 12,62 | 20,30 | 36,39 | 59,23 | 84,46 | 112,83 |
| | New (sec) | 0,63 | 2,38 | 6,65 | 12,00 | 18,89 | 30,20 | 31,23 | 58,10 |
| | Train Savings | 42,86 | 50,16 | 49,83 | 52,02 | 54,35 | 48,68 | 60,62 | 48,79 |
| soybean | Old (sec) | 0,38 | 1,71 | 3,12 | 5,17 | 8,54 | 12,84 | 17,91 | 22,29 |
| | New (sec) | 0,25 | 1,51 | 2,53 | 4,59 | 7,48 | 11,59 | 14,72 | 19,01 |
| | Train Savings | 46,98 | 56,04 | 45,00 | 46,84 | 45,44 | 46,70 | 43,65 | 43,85 |
| vote | Old (sec) | 0,81 | 3,30 | 7,24 | 12,83 | 21,33 | 28,66 | 41,91 | 54,05 |
| | New (sec) | 0,60 | 2,47 | 5,47 | 9,67 | 15,72 | 21,65 | 30,01 | 40,15 |
| | Train Savings | 27,88 | 29,93 | 29,59 | 29,55 | 30,11 | 29,30 | 29,64 | 29,37 |
| anneal | Old (sec) | 5,89 | 18,45 | 55,60 | 98,84 | 155,75 | 165,55 | 308,79 | 368,76 |
| | New (sec) | 3,46 | 12,71 | 34,06 | 59,47 | 102,35 | 133,83 | 165,07 | 161,76 |
| | Train Savings | 36,88 | 34,19 | 33,89 | 34,50 | 34,40 | 31,68 | 33,89 | 57,62 |
| crime | Old (sec) | 0,70 | 4,03 | 9,13 | 14,42 | 21,61 | 32,52 | 55,32 | 61,77 |
| | New (sec) | 0,60 | 2,43 | 6,54 | 11,03 | 17,16 | 27,65 | 32,39 | 47,82 |
| | Train Savings | 28,65 | 43,53 | 33,45 | 48,77 | 30,03 | 38,32 | 53,43 | 41,42 |
| kr-vs-kp | Old (sec) | 5,09 | 19,46 | 43,86 | 76,66 | 154,44 | 193,63 | 225,69 | 400,59 |
| | New (sec) | 3,82 | 11,49 | 28,37 | 58,19 | 105,02 | 143,92 | 157,79 | 235,42 |
| | Train Savings | 27,26 | 54,45 | 49,76 | 29,13 | 28,93 | 28,96 | 49,92 | 28,89 |
| breast | Old (sec) | 0,67 | 3,79 | 8,50 | 17,50 | 24,28 | 37,39 | 55,97 | 65,66 |
| | New (sec) | 0,41 | 1,80 | 4,30 | 8,17 | 11,49 | 21,83 | 28,02 | 37,08 |
| | Train Savings | 42,81 | 53,88 | 53,69 | 55,37 | 53,54 | 58,83 | 63,45 | 61,62 |
| multiplexor | Old (sec) | 0,36 | 1,66 | 4,12 | 8,91 | 16,23 | 9,62 | 30,47 | 34,34 |
| | New (sec) | 0,28 | 1,46 | 3,34 | 7,20 | 6,17 | 9,18 | 30,15 | 27,11 |
| | Train Savings | 50,59 | 50,99 | 56,33 | 65,01 | 42,22 | 41,54 | 66,33 | 63,62 |

Table 7. A savings-only comparison with mutation rate at 0.01 and $x$ in $[10^4..10^5]$

| Dataset | | 100 | 200 | 300 | 400 | 500 | 600 | 700 | 800 |
|---|---|---|---|---|---|---|---|---|---|
| balance-scale | Old (sec) | 0,96 | 4,28 | 12,36 | 20,17 | 31,52 | 53,72 | 69,84 | 91,78 |
| | New (sec) | 0,49 | 2,15 | 7,48 | 10,53 | 16,27 | 26,59 | 36,19 | 45,58 |
| | Train Savings | 64,30 | 65,64 | 78,74 | 71,53 | 73,59 | 77,70 | 78,71 | 76,35 |
| zoo | Old (sec) | 0,41 | 1,68 | 5,49 | 9,86 | 15,47 | 19,72 | 34,39 | 35,86 |
| | New (sec) | 0,26 | 1,11 | 3,37 | 6,90 | 11,01 | 14,25 | 29,97 | 24,73 |
| | Train Savings | 57,20 | 65,07 | 69,42 | 71,88 | 73,12 | 65,71 | 75,09 | 64,49 |
| credit-a | Old (sec) | 1,54 | 7,04 | 12,49 | 28,06 | 48,88 | 71,52 | 73,51 | 135,54 |
| | New (sec) | 0,68 | 3,01 | 5,99 | 11,92 | 22,57 | 31,92 | 35,67 | 51,59 |
| | Train Savings | 57,61 | 59,03 | 55,35 | 64,76 | 54,10 | 59,13 | 55,28 | 69,33 |
| lymph | Old (sec) | 0,45 | 3,04 | 5,63 | 9,79 | 24,64 | 47,44 | 47,94 | 62,53 |
| | New (sec) | 0,27 | 1,96 | 3,25 | 5,95 | 15,69 | 29,62 | 28,22 | 38,01 |
| | Train Savings | 51,11 | 70,08 | 69,58 | 61,71 | 76,36 | 72,71 | 78,30 | 76,03 |
| glass | Old (sec) | 0,60 | 5,19 | 12,25 | 24,84 | 37,41 | 107,83 | 113,69 | 127,55 |
| | New (sec) | 0,26 | 2,23 | 4,94 | 15,81 | 17,97 | 37,30 | 41,54 | 51,21 |
| | Train Savings | 61,91 | 60,32 | 61,09 | 59,12 | 76,90 | 58,15 | 69,03 | 72,50 |
| soybean | Old (sec) | 0,32 | 0,96 | 3,91 | 5,91 | 9,51 | 12,48 | 18,08 | 27,78 |
| | New (sec) | 0,23 | 0,70 | 2,86 | 4,51 | 7,10 | 9,24 | 13,96 | 20,69 |
| | Train Savings | 58,62 | 50,99 | 67,19 | 59,77 | 59,38 | 61,43 | 58,78 | 64,20 |
| vote | Old (sec) | 0,81 | 3,35 | 7,39 | 13,04 | 28,18 | 44,91 | 39,85 | 53,30 |
| | New (sec) | 0,47 | 1,98 | 4,45 | 8,22 | 12,83 | 19,61 | 24,48 | 32,47 |
| | Train Savings | 42,98 | 43,27 | 43,25 | 43,33 | 42,59 | 77,49 | 42,68 | 42,67 |
| anneal | Old (sec) | 4,83 | 19,29 | 41,87 | 102,74 | 143,96 | 166,60 | 289,42 | 450,48 |
| | New (sec) | 2,62 | 10,35 | 23,53 | 51,03 | 65,07 | 94,48 | 149,41 | 201,70 |
| | Train Savings | 44,32 | 45,52 | 42,75 | 46,89 | 43,35 | 42,72 | 47,70 | 47,91 |
| crime | Old (sec) | 0,68 | 2,86 | 10,72 | 15,28 | 25,98 | 33,56 | 43,24 | 80,87 |
| | New (sec) | 0,49 | 2,04 | 5,42 | 10,09 | 13,37 | 23,00 | 31,68 | 51,91 |
| | Train Savings | 41,76 | 41,68 | 65,85 | 44,91 | 68,04 | 45,57 | 44,40 | 62,72 |
| kr-vs-kp | Old (sec) | 5,07 | 20,31 | 46,21 | 82,32 | 133,11 | 196,38 | 448,10 | 398,74 |
| | New (sec) | 2,82 | 11,86 | 25,75 | 46,22 | 86,26 | 107,87 | 118,28 | 180,32 |
| | Train Savings | 42,11 | 41,64 | 42,13 | 42,82 | 47,14 | 42,24 | 64,71 | 42,58 |
| breast | Old (sec) | 1,40 | 3,30 | 10,90 | 16,10 | 26,44 | 39,90 | 73,23 | 63,57 |
| | New (sec) | 0,41 | 1,24 | 4,04 | 6,72 | 10,00 | 19,60 | 30,10 | 27,89 |
| | Train Savings | 71,25 | 59,99 | 70,79 | 67,69 | 74,72 | 74,31 | 75,30 | 71,16 |
| multiplexor | Old (sec) | 0,31 | 0,97 | 3,23 | 15,46 | 16,29 | 23,48 | 28,93 | 16,27 |
| | New (sec) | 0,23 | 0,72 | 2,55 | 9,84 | 14,02 | 25,99 | 27,82 | 13,24 |
| | Train Savings | 63,38 | 56,59 | 65,54 | 79,59 | 77,41 | 81,38 | 79,56 | 55,73 |

The key observation, of course, is that these real datasets deliver sizeable and consistent savings. The consistency is observed across larger experiments and within how instance test savings are translated into actual elapsed training time. The latter is of paramount importance; not only does it suggest that the housekeeping operations seem to be contained (even in a rather straightforward proof-of-concept implementation) but also that the approximation strategy we used with the two bounding cases to estimate the benefits is robust enough to apply to practical in-between problems equally well.

The careful reader might note here that our theoretical improvement estimate was based on the concept of a *node-instance-check* as a unit of improvement and that results above report instance savings (along a path whose length is not taken into account). We acknowledge the simplification; still the reduced resolution of the results we report is more than satisfactory in terms of appreciating the new potential.

For the sake of completeness, we now report some accuracy-size-time results on the above experimental configurations as compared to J48, a C4.5 clone that is available within the WEKA package (Witten & Frank, 2005). The only extension is that to test accuracy we used a 5-fold cross-validation; we then averaged the size and the time as well across each cross-validation run. To avoid cluttering we omit standard deviations. Moreover, we used the default parameters for J48 and we opted to generate binary decision trees and report the results for both its un-pruned and pruned versions – the latter shown as J48(p). The results are shown as follows (for GATree we only report the "new" version):

- For a mutation rate of 0.5
    - the *x*-factor was set at $10^4$ (see Table 8),
    - the *x*-factor was set at $10^4$ at the start of the evolution and increased linearly up to $10^5$ at the end of the evolution (see Table 9).
- For a mutation rate of 0.01
    - the *x*-factor was set at $10^4$ (see Table 10),
    - the *x*-factor was set at $10^4$ at the start of the evolution and increased linearly up to $10^5$ at the end of the evolution (see Table 11).

For the sake of completeness, we now report some accuracy-size-time results on the above experimental configurations as compared to J48, a C4.5 clone that is available within the WEKA package (Witten & Frank, 2005). The only extension is that to test accuracy we used a 5-fold cross-validation; we then averaged the size and the time as well across each cross-validation run. To avoid cluttering we omit standard deviations. Moreover, we used the default parameters for J48 and we opted to generate binary decision trees and report the results for both its un-pruned and pruned versions – the latter shown as J48(p). The results are shown as follows (for GATree we only report the "new" version; furthermore note that accuracy is reported in the [0..1] range):

Table 8. An accuracy-size-time comparison with mutation rate at 0.5 and $x$ at $10^4$

| Dataset | | 100 | 200 | 300 | 400 | 500 | 600 | 700 | 800 | J48 | J48 (p) |
|---|---|---|---|---|---|---|---|---|---|---|---|
| balance-scale | Accuracy | 0,690 | 0,686 | 0,704 | 0,704 | 0,744 | 0,702 | 0,715 | 0,690 | 0,776 | 0,781 |
| | Size | 14 | 13 | 15 | 13 | 17 | 16 | 14 | 16 | 103 | 55 |
| | Time | 0,65 | 3,67 | 7,26 | 12,76 | 19,71 | 27,04 | 39,88 | 54,31 | 0,06 | 0,04 |
| zoo | Accuracy | 0,840 | 0,870 | 0,860 | 0,890 | 0,900 | 0,940 | 0,920 | 0,900 | 0,911 | 0,881 |
| | Size | 9 | 11 | 11 | 13 | 13 | 14 | 16 | 13 | 17 | 13 |
| | Time | 0,74 | 3,89 | 12,55 | 15,39 | 20,66 | 30,90 | 37,45 | 45,10 | 0,01 | 0,01 |
| credit-a | Accuracy | 0,848 | 0,846 | 0,849 | 0,857 | 0,845 | 0,845 | 0,842 | 0,851 | 0,830 | 0,848 |
| | Size | 5 | 8 | 9 | 10 | 10 | 9 | 9 | 9 | 49 | 23 |
| | Time | 0,80 | 3,56 | 8,04 | 13,84 | 20,66 | 30,03 | 41,49 | 55,00 | 0,09 | 0,08 |
| lymph | Accuracy | 0,759 | 0,738 | 0,766 | 0,731 | 0,766 | 0,779 | 0,779 | 0,772 | 0,770 | 0,764 |
| | Size | 15 | 11 | 13 | 15 | 19 | 20 | 17 | 18 | 27 | 13 |
| | Time | 0,57 | 2,53 | 6,26 | 11,12 | 18,62 | 28,07 | 37,56 | 46,93 | 0,03 | 0,01 |
| glass | Accuracy | 0,524 | 0,524 | 0,495 | 0,538 | 0,567 | 0,543 | 0,595 | 0,581 | 0,486 | 0,533 |
| | Size | 15 | 16 | 19 | 20 | 21 | 21 | 20 | 24 | 57 | 39 |
| | Time | 0,96 | 4,59 | 11,47 | 19,67 | 28,86 | 45,53 | 61,86 | 89,64 | 0,09 | 0,01 |
| soybean | Accuracy | 0,971 | 0,943 | 0,971 | 0,914 | 0,971 | 0,971 | 1,000 | 0,971 | 0,949 | 0,923 |
| | Size | 7 | 7 | 7 | 7 | 7 | 7 | 7 | 7 | 7 | 7 |
| | Time | 0,40 | 1,58 | 3,64 | 6,25 | 9,69 | 14,00 | 19,70 | 25,06 | 0,01 | 0,01 |
| vote | Accuracy | 0,956 | 0,956 | 0,956 | 0,956 | 0,956 | 0,956 | 0,956 | 0,956 | 0,966 | 0,954 |
| | Size | 3 | 4 | 3 | 3 | 3 | 3 | 3 | 3 | 11 | 9 |
| | Time | 0,63 | 2,75 | 7,05 | 12,84 | 20,84 | 29,49 | 41,39 | 54,10 | 0,01 | 0,01 |
| anneal | Accuracy | 0,817 | 0,837 | 0,838 | 0,867 | 0,842 | 0,883 | 0,868 | 0,902 | 0,954 | 0,938 |
| | Size | 5 | 5 | 4 | 9 | 6 | 9 | 8 | 9 | 49 | 35 |
| | Time | 2,17 | 9,30 | 20,86 | 37,38 | 59,68 | 81,55 | 115,22 | 159,42 | 0,08 | 0,11 |
| crime | Accuracy | 0,712 | 0,715 | 0,724 | 0,744 | 0,730 | 0,721 | 0,721 | 0,739 | 0,728 | 0,728 |
| | Size | 5 | 7 | 10 | 11 | 13 | 13 | 10 | 12 | 65 | 27 |
| | Time | 0,67 | 2,99 | 6,81 | 12,36 | 20,01 | 29,22 | 39,26 | 49,11 | 0,09 | 0,13 |
| kr-vs-kp | Accuracy | 0,912 | 0,932 | 0,924 | 0,948 | 0,969 | 0,965 | 0,976 | 0,970 | 0,993 | 0,989 |
| | Size | 4 | 6 | 4 | 8 | 11 | 12 | 13 | 11 | 57 | 51 |
| | Time | 2,47 | 10,28 | 24,23 | 38,03 | 57,80 | 76,42 | 106,45 | 132,52 | 0,11 | 0,04 |
| breast | Accuracy | 0,737 | 0,730 | 0,737 | 0,712 | 0,730 | 0,709 | 0,709 | 0,740 | 0,710 | 0,664 |
| | Size | 8 | 9 | 13 | 15 | 14 | 16 | 15 | 17 | 43 | 39 |
| | Time | 0,59 | 2,57 | 6,22 | 11,20 | 18,01 | 27,90 | 37,73 | 49,64 | 0,06 | 0,03 |
| multiplexor | Accuracy | 0,540 | 0,690 | 0,610 | 0,720 | 0,670 | 0,660 | 0,670 | 0,680 | 0,630 | 0,550 |
| | Size | 11 | 12 | 17 | 16 | 16 | 23 | 23 | 21 | 23 | 21 |
| | Time | 0,31 | 1,56 | 4,04 | 8,10 | 11,89 | 20,19 | 28,03 | 38,25 | 0,01 | 0,01 |

Table 9. An accuracy-size-time comparison with mutation rate at 0.5 and $x$ in $[10^4..10^5]$

| Dataset | | 100 | 200 | 300 | 400 | 500 | 600 | 700 | 800 | J48 | J48 (p) |
|---|---|---|---|---|---|---|---|---|---|---|---|
| balance-scale | Accuracy | 0,709 | 0,702 | 0,733 | 0,733 | 0,742 | 0,744 | 0,771 | 0,736 | 0,776 | 0,781 |
| | Size | 20 | 26 | 32 | 30 | 33 | 30 | 26 | 31 | 103 | 55 |
| | Time | 0,77 | 3,82 | 8,89 | 16,40 | 26,58 | 39,33 | 50,70 | 83,69 | 0,06 | 0,04 |
| zoo | Accuracy | 0,820 | 0,890 | 0,890 | 0,920 | 0,920 | 0,940 | 0,940 | 0,950 | 0,911 | 0,881 |
| | Size | 20 | 11 | 13 | 15 | 15 | 19 | 18 | 16 | 17 | 13 |
| | Time | 1,36 | 5,07 | 6,88 | 14,06 | 20,83 | 31,69 | 39,66 | 54,10 | 0,01 | 0,01 |
| credit-a | Accuracy | 0,848 | 0,841 | 0,854 | 0,838 | 0,848 | 0,851 | 0,848 | 0,849 | 0,830 | 0,848 |
| | Size | 11 | 17 | 17 | 17 | 18 | 20 | 19 | 25 | 49 | 23 |
| | Time | 0,87 | 3,73 | 8,28 | 15,29 | 28,22 | 37,13 | 52,57 | 69,69 | 0,09 | 0,08 |
| lymph | Accuracy | 0,772 | 0,779 | 0,793 | 0,766 | 0,752 | 0,759 | 0,807 | 0,800 | 0,770 | 0,764 |
| | Size | 26 | 28 | 28 | 21 | 30 | 31 | 24 | 34 | 27 | 13 |
| | Time | 0,81 | 3,99 | 8,65 | 15,20 | 28,26 | 39,97 | 50,13 | 78,97 | 0,03 | 0,01 |
| glass | Accuracy | 0,519 | 0,557 | 0,576 | 0,533 | 0,567 | 0,529 | 0,538 | 0,567 | 0,486 | 0,533 |
| | Size | 37 | 34 | 46 | 45 | 41 | 46 | 44 | 44 | 57 | 39 |
| | Time | 1,26 | 5,92 | 17,53 | 32,96 | 46,53 | 85,77 | 109,59 | 213,48 | 0,09 | 0,01 |
| soybean | Accuracy | 0,886 | 0,971 | 0,943 | 0,971 | 1,000 | 0,943 | 0,943 | 0,943 | 0,949 | 0,923 |
| | Size | 7 | 8 | 7 | 7 | 7 | 7 | 7 | 7 | 7 | 7 |
| | Time | 0,53 | 1,90 | 4,53 | 6,09 | 10,20 | 14,24 | 19,37 | 25,54 | 0,01 | 0,01 |
| vote | Accuracy | 0,956 | 0,959 | 0,956 | 0,947 | 0,956 | 0,956 | 0,952 | 0,954 | 0,966 | 0,954 |
| | Size | 11 | 5 | 7 | 5 | 9 | 9 | 8 | 7 | 11 | 9 |
| | Time | 0,67 | 3,02 | 6,89 | 13,26 | 17,94 | 27,49 | 41,63 | 59,10 | 0,01 | 0,01 |
| anneal | Accuracy | 0,811 | 0,853 | 0,894 | 0,865 | 0,895 | 0,879 | 0,926 | 0,932 | 0,954 | 0,938 |
| | Size | 11 | 6 | 11 | 15 | 13 | 15 | 16 | 17 | 49 | 35 |
| | Time | 2,22 | 9,08 | 19,59 | 38,40 | 58,11 | 84,36 | 121,11 | 154,04 | 0,08 | 0,11 |
| crime | Accuracy | 0,708 | 0,712 | 0,715 | 0,735 | 0,717 | 0,726 | 0,717 | 0,739 | 0,728 | 0,728 |
| | Size | 10 | 11 | 27 | 28 | 29 | 22 | 26 | 31 | 65 | 27 |
| | Time | 0,79 | 2,76 | 8,02 | 16,92 | 25,28 | 36,44 | 52,76 | 69,65 | 0,09 | 0,13 |
| kr-vs-kp | Accuracy | 0,920 | 0,916 | 0,941 | 0,940 | 0,955 | 0,975 | 0,974 | 0,982 | 0,993 | 0,989 |
| | Size | 4 | 6 | 11 | 9 | 16 | 18 | 19 | 21 | 57 | 51 |
| | Time | 2,34 | 9,62 | 21,67 | 40,88 | 58,85 | 86,04 | 95,52 | 143,86 | 0,11 | 0,04 |
| breast | Accuracy | 0,733 | 0,733 | 0,719 | 0,730 | 0,705 | 0,695 | 0,716 | 0,688 | 0,710 | 0,664 |
| | Size | 16 | 27 | 31 | 31 | 39 | 31 | 33 | 33 | 43 | 39 |
| | Time | 0,65 | 3,58 | 8,88 | 15,23 | 27,30 | 39,06 | 54,69 | 75,76 | 0,06 | 0,03 |
| multiplexor | Accuracy | 0,580 | 0,690 | 0,620 | 0,610 | 0,640 | 0,640 | 0,690 | 0,710 | 0,630 | 0,550 |
| | Size | 12 | 17 | 33 | 23 | 27 | 39 | 39 | 39 | 23 | 21 |
| | Time | 0,39 | 3,01 | 9,22 | 13,01 | 24,45 | 36,64 | 52,33 | 69,42 | 0,01 | 0,01 |

Table 10. An accuracy-size-time comparison with mutation rate at 0.01 and $x$ at $10^4$

| Dataset | | 100 | 200 | 300 | 400 | 500 | 600 | 700 | 800 | J48 | J48 (p) |
|---|---|---|---|---|---|---|---|---|---|---|---|
| balance-scale | Accuracy | 0,667 | 0,686 | 0,672 | 0,707 | 0,693 | 0,677 | 0,682 | 0,678 | 0,776 | 0,781 |
| | Size | 11 | 12 | 12 | 13 | 12 | 13 | 12 | 11 | 103 | 55 |
| | Time | 0,51 | 2,01 | 4,72 | 8,85 | 13,40 | 20,60 | 27,75 | 34,59 | 0,06 | 0,04 |
| zoo | Accuracy | 0,710 | 0,800 | 0,840 | 0,840 | 0,850 | 0,890 | 0,860 | 0,860 | 0,911 | 0,881 |
| | Size | 7 | 8 | 9 | 8 | 8 | 10 | 9 | 11 | 17 | 13 |
| | Time | 0,37 | 1,56 | 3,82 | 7,00 | 10,77 | 15,98 | 22,07 | 30,45 | 0,01 | 0,01 |
| credit-a | Accuracy | 0,852 | 0,855 | 0,855 | 0,843 | 0,859 | 0,854 | 0,851 | 0,849 | 0,830 | 0,848 |
| | Size | 3 | 3 | 3 | 6 | 4 | 8 | 6 | 7 | 49 | 23 |
| | Time | 0,55 | 2,25 | 5,10 | 9,74 | 14,34 | 21,17 | 29,24 | 41,45 | 0,09 | 0,08 |
| lymph | Accuracy | 0,690 | 0,766 | 0,752 | 0,724 | 0,738 | 0,779 | 0,752 | 0,752 | 0,770 | 0,764 |
| | Size | 7 | 6 | 9 | 9 | 11 | 12 | 11 | 13 | 27 | 13 |
| | Time | 0,44 | 1,89 | 4,56 | 8,17 | 13,07 | 21,03 | 25,99 | 34,74 | 0,03 | 0,01 |
| glass | Accuracy | 0,490 | 0,476 | 0,429 | 0,505 | 0,543 | 0,519 | 0,519 | 0,557 | 0,486 | 0,533 |
| | Size | 10 | 9 | 13 | 10 | 12 | 11 | 13 | 14 | 57 | 39 |
| | Time | 0,73 | 2,89 | 7,04 | 12,53 | 20,64 | 28,92 | 41,65 | 62,96 | 0,09 | 0,01 |
| soybean | Accuracy | 0,857 | 0,886 | 1,000 | 1,000 | 0,943 | 0,943 | 0,971 | 0,971 | 0,949 | 0,923 |
| | Size | 6 | 7 | 7 | 8 | 7 | 7 | 7 | 7 | 7 | 7 |
| | Time | 0,29 | 1,27 | 3,03 | 5,57 | 8,36 | 11,75 | 16,84 | 21,54 | 0,01 | 0,01 |
| vote | Accuracy | 0,952 | 0,952 | 0,954 | 0,956 | 0,956 | 0,956 | 0,956 | 0,956 | 0,966 | 0,954 |
| | Size | 3 | 3 | 3 | 3 | 3 | 3 | 3 | 3 | 11 | 9 |
| | Time | 0,50 | 2,08 | 4,70 | 8,36 | 13,49 | 18,56 | 26,81 | 33,90 | 0,01 | 0,01 |
| anneal | Accuracy | 0,790 | 0,823 | 0,838 | 0,815 | 0,859 | 0,837 | 0,849 | 0,842 | 0,954 | 0,938 |
| | Size | 5 | 5 | 3 | 7 | 5 | 5 | 5 | 6 | 49 | 35 |
| | Time | 1,74 | 7,46 | 15,20 | 30,81 | 39,16 | 64,20 | 85,56 | 103,42 | 0,08 | 0,11 |
| crime | Accuracy | 0,723 | 0,715 | 0,719 | 0,717 | 0,719 | 0,710 | 0,721 | 0,715 | 0,728 | 0,728 |
| | Size | 3 | 5 | 4 | 5 | 5 | 5 | 5 | 7 | 65 | 27 |
| | Time | 0,52 | 2,12 | 5,19 | 9,67 | 13,97 | 21,17 | 30,50 | 37,83 | 0,09 | 0,13 |
| kr-vs-kp | Accuracy | 0,908 | 0,918 | 0,924 | 0,927 | 0,920 | 0,927 | 0,924 | 0,930 | 0,993 | 0,989 |
| | Size | 3 | 4 | 4 | 5 | 3 | 5 | 4 | 4 | 57 | 51 |
| | Time | 1,93 | 7,21 | 17,53 | 32,80 | 50,29 | 64,26 | 92,82 | 125,62 | 0,11 | 0,04 |
| breast | Accuracy | 0,709 | 0,712 | 0,751 | 0,740 | 0,723 | 0,719 | 0,719 | 0,712 | 0,710 | 0,664 |
| | Size | 5 | 7 | 7 | 8 | 8 | 9 | 9 | 9 | 43 | 39 |
| | Time | 0,45 | 1,87 | 4,45 | 7,81 | 11,76 | 18,76 | 24,70 | 31,95 | 0,06 | 0,03 |
| multiplexor | Accuracy | 0,600 | 0,550 | 0,600 | 0,570 | 0,680 | 0,660 | 0,600 | 0,650 | 0,630 | 0,550 |
| | Size | 5 | 6 | 11 | 5 | 14 | 12 | 10 | 12 | 23 | 21 |
| | Time | 0,22 | 1,04 | 3,17 | 4,18 | 9,57 | 13,21 | 17,39 | 25,07 | 0,01 | 0,01 |

Table 11. An accuracy-size-time comparison with mutation rate at 0.01 and $x$ in $[10^4..10^5]$

| Dataset | | 100 | 200 | 300 | 400 | 500 | 600 | 700 | 800 | J48 | J48 (p) |
|---|---|---|---|---|---|---|---|---|---|---|---|
| balance-scale | Accuracy | 0,661 | 0,706 | 0,659 | 0,682 | 0,659 | 0,744 | 0,702 | 0,731 | 0,776 | 0,781 |
| | Size | 10 | 19 | 15 | 20 | 20 | 25 | 23 | 23 | 103 | 55 |
| | Time | 0,56 | 2,83 | 5,90 | 11,78 | 798,54 | 29,75 | 39,98 | 55,20 | 0,06 | 0,04 |
| zoo | Accuracy | 0,660 | 0,820 | 0,830 | 0,820 | 0,860 | 0,860 | 0,860 | 0,870 | 0,911 | 0,881 |
| | Size | 9 | 8 | 8 | 8 | 8 | 10 | 11 | 10 | 17 | 13 |
| | Time | 0,47 | 1,99 | 4,62 | 7,12 | 894,54 | 18,05 | 24,95 | 31,88 | 0,01 | 0,01 |
| credit-a | Accuracy | 0,855 | 0,855 | 0,852 | 0,857 | 0,854 | 0,852 | 0,855 | 0,846 | 0,830 | 0,848 |
| | Size | 3 | 5 | 4 | 7 | 7 | 9 | 7 | 9 | 49 | 23 |
| | Time | 0,56 | 2,40 | 5,40 | 10,18 | 16,31 | 26,31 | 31,86 | 47,81 | 0,09 | 0,08 |
| lymph | Accuracy | 0,717 | 0,759 | 0,779 | 0,793 | 0,793 | 0,772 | 0,772 | 0,800 | 0,770 | 0,764 |
| | Size | 8 | 15 | 15 | 23 | 14 | 16 | 16 | 17 | 27 | 13 |
| | Time | 0,52 | 2,70 | 6,09 | 12,59 | 16,22 | 24,18 | 35,83 | 47,17 | 0,03 | 0,01 |
| glass | Accuracy | 0,443 | 0,457 | 0,519 | 0,486 | 0,471 | 0,567 | 0,543 | 0,586 | 0,486 | 0,533 |
| | Size | 6 | 8 | 16 | 15 | 15 | 18 | 18 | 24 | 57 | 39 |
| | Time | 0,66 | 3,29 | 9,48 | 16,42 | 24,51 | 38,11 | 49,34 | 71,70 | 0,09 | 0,01 |
| soybean | Accuracy | 0,857 | 0,914 | 1,000 | 0,943 | 1,000 | 1,000 | 1,000 | 0,943 | 0,949 | 0,923 |
| | Size | 7 | 7 | 8 | 7 | 7 | 7 | 7 | 7 | 7 | 7 |
| | Time | 0,35 | 1,41 | 3,41 | 5,86 | 9,10 | 12,65 | 17,18 | 22,57 | 0,01 | 0,01 |
| vote | Accuracy | 0,952 | 0,949 | 0,956 | 0,952 | 0,952 | 0,956 | 0,956 | 0,956 | 0,966 | 0,954 |
| | Size | 3 | 4 | 5 | 7 | 4 | 3 | 3 | 5 | 11 | 9 |
| | Time | 0,52 | 2,19 | 4,95 | 9,20 | 14,04 | 19,26 | 26,61 | 384,86 | 0,01 | 0,01 |
| anneal | Accuracy | 0,785 | 0,813 | 0,832 | 0,836 | 0,856 | 0,849 | 0,846 | 0,863 | 0,954 | 0,938 |
| | Size | 4 | 6 | 4 | 4 | 4 | 7 | 7 | 7 | 49 | 35 |
| | Time | 1,83 | 7,52 | 16,69 | 33,04 | 40,16 | 70,28 | 98,83 | 125,17 | 0,08 | 0,11 |
| crime | Accuracy | 0,723 | 0,721 | 0,715 | 0,723 | 0,726 | 0,708 | 0,701 | 0,706 | 0,728 | 0,728 |
| | Size | 5 | 4 | 7 | 8 | 11 | 13 | 8 | 12 | 65 | 27 |
| | Time | 0,57 | 2,31 | 5,40 | 11,16 | 17,43 | 26,19 | 31,60 | 46,58 | 0,09 | 0,13 |
| kr-vs-kp | Accuracy | 0,904 | 0,918 | 0,919 | 0,920 | 0,937 | 0,934 | 0,917 | 0,932 | 0,993 | 0,989 |
| | Size | 4 | 5 | 3 | 5 | 8 | 7 | 3 | 7 | 57 | 51 |
| | Time | 1,92 | 7,65 | 17,29 | 29,75 | 42,65 | 64,13 | 95,19 | 110,66 | 0,11 | 0,04 |
| breast | Accuracy | 0,737 | 0,726 | 0,712 | 0,719 | 0,705 | 0,719 | 0,705 | 0,712 | 0,710 | 0,664 |
| | Size | 6 | 8 | 13 | 11 | 16 | 13 | 22 | 20 | 43 | 39 |
| | Time | 0,49 | 2,01 | 5,56 | 8,77 | 16,93 | 22,19 | 36,64 | 52,29 | 0,06 | 0,03 |
| multiplexor | Accuracy | 0,570 | 0,640 | 0,610 | 0,620 | 0,650 | 0,680 | 0,690 | 0,650 | 0,630 | 0,550 |
| | Size | 6 | 9 | 12 | 9 | 13 | 13 | 17 | 16 | 23 | 21 |
| | Time | 0,29 | 1,50 | 3,69 | 5,98 | 11,39 | 17,26 | 29,44 | 35,50 | 0,01 | 0,01 |

The above indicative results show that for most data sets GATree can deliver a very sizeable size benefit, even compared to J48(p), for the extra time it spends exploring the population. Accuracy is overall comparable to J48(p), if not better, but there are a few accuracy glitches in the data-sets *balance-scale*, *anneal* and *kr-vs-kp*, which are due to the fact that these data sets are (J48-wise) better served by larger trees, which are not rewarded by GATree with its current *x*-factor settings. A very clear hint that this is the reason for this behavior is that GATree accuracy increases consistently with time, since larger trees are more rewarded towards the end of the evolution (whereas, one can note, the other data sets demonstrate a relative stabilization in accuracy very early in the evolution).

One could opt to speed-up the updating of the x-factor; in that case one would obtain good accuracy results for *balance-scale*, *anneal* and *kr-vs-kp* considerably faster. A limited-scope experiment is shown in Table 12, where one can observe that behavior.

Table 12. An accuracy-size-time comparison with mutation rate at 0.01 and *x* in $[10^4..10^6]$

| Dataset | | Mutation 50% | | | | | | J48 | J48 (p) |
|---|---|---|---|---|---|---|---|---|---|
| | | 100 | 200 | 300 | 400 | 500 | 600 | | |
| balance-scale | Accuracy | 0,762 | 0,683 | 0,715 | 0,715 | 0,773 | 0,755 | 0,776 | 0,781 |
| | Size | 52 | 34 | 52 | 66 | 48 | 80 | 103 | 55 |
| | Time | 0,92 | 5,27 | 15,33 | 31,41 | 44,78 | 98,00 | 0,06 | 0,04 |
| anneal | Accuracy | 0,817 | 0,847 | 0,868 | 0,892 | 0,911 | 0,927 | 0,954 | 0,938 |
| | Size | 15 | 7 | 13 | 21 | 20 | 22 | 49 | 35 |
| | Time | 2,39 | 8,89 | 20,84 | 37,79 | 62,82 | 89,74 | 0,08 | 0,11 |
| kr-vs-kp | Accuracy | 0,939 | 0,950 | 0,967 | 0,951 | 0,982 | 0,961 | 0,993 | 0,989 |
| | Size | 22 | 15 | 19 | 11 | 30 | 21 | 57 | 51 |
| | Time | 2,24 | 7,82 | 16,49 | 33,23 | 54,03 | 76,06 | 0,11 | 0,04 |

| Dataset | | Mutation 1% | | | | | | J48 | J48 (p) |
|---|---|---|---|---|---|---|---|---|---|
| | | 100 | 200 | 300 | 400 | 500 | 600 | | |
| balance-scale | Accuracy | 0,680 | 0,680 | 0,707 | 0,733 | 0,718 | 0,720 | 0,776 | 0,781 |
| | Size | 11 | 17 | 30 | 36 | 35 | 45 | 103 | 55 |
| | Time | 0,57 | 2,95 | 10,32 | 19,74 | 30,94 | 59,08 | 0,06 | 0,04 |
| anneal | Accuracy | 0,779 | 0,820 | 0,823 | 0,855 | 0,849 | 0,838 | 0,954 | 0,938 |
| | Size | 3 | 5 | 5 | 7 | 7 | 9 | 49 | 35 |
| | Time | 1,63 | 7,01 | 16,12 | 27,15 | 850,58 | 64,73 | 0,08 | 0,11 |
| kr-vs-kp | Accuracy | 0,915 | 0,919 | 0,912 | 0,924 | 0,934 | 0,931 | 0,993 | 0,989 |
| | Size | 4 | 5 | 4 | 6 | 9 | 5 | 57 | 51 |
| | Time | 1,73 | 6,81 | 14,60 | 25,56 | 39,87 | 56,91 | 0,11 | 0,04 |

Incidentally, we note that decision tree evolution seems to be a problem that is better served by larger mutation rates. Of course, one might decide to first generate a population of decision trees (via J48, for example) and, subsequently, use GATree with a small *x*-factor to bias evolutionary refinement towards smaller trees; in such case it is rather likely that we would need a small mutation value, as is usual in most evolution problems.

The above exploratory approach to model development is facilitated by GATree, since it allows observing how the accuracy and the size of a model develop during evolution. For example, one can note (see Figure 6) how quite early on it is possible to have a good model at a small size by watching a 5-fold cross-validation on the *multiplexor* problem. Therein, drops in accuracy refer to each new run of the cross-validation scheme and accuracy then increases towards the end of each run; also note how the spike in model size (at about the middle of the evolution) corresponds to a relatively small accuracy in that run.

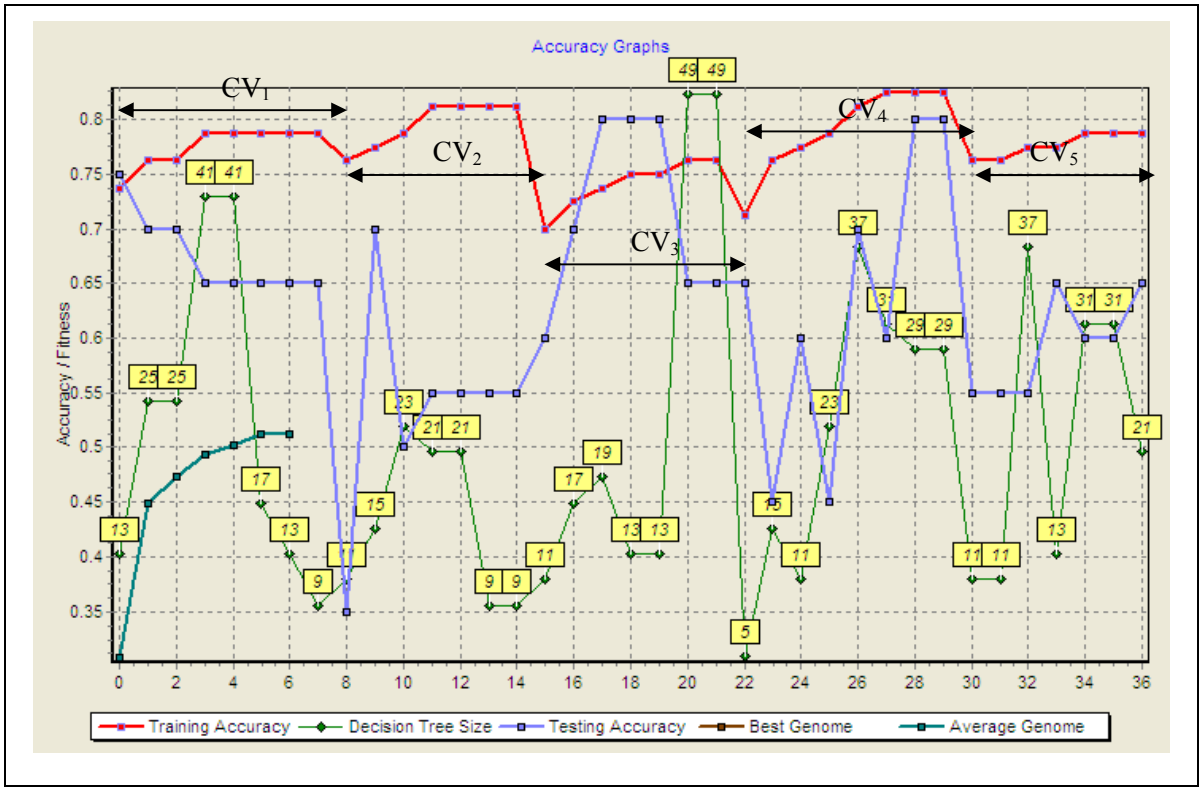

Figure 6: An example of exploratory evolution.

To further delve into accuracy and size aspects, to research the possibilities for dynamically tuning the *x*-factor and the effect this may have on some data sets (of theoretical or practical importance), as well as the on-the-fly modifications of the evolution parameters and the associated constraints, we refer the interested reader to the experimental results and the discussion in previous work (Papagelis & Kalles, 2001; Kalles & Pierrakeas, 2006).

## 6. Conclusions and future directions

We have presented an extension to a genetic algorithm that evolves binary decision trees to improve the evolution speed by reusing past fitness calculations. Such reuse is possible by observing that the divide-and-conquer formulation of the fitness function allows for its lossless piecewise calculation. We have analyzed two bounding cases (of very thin and very bushy trees) to estimate the range of possible speed-ups and we have discussed the validity of our assumptions and our results vis-à-vis the potential for using the same method in other domains, beyond decision trees. We have backed our claims with an indicative yet extensive experimental evaluation to show that savings are substantial and across a diverse number of datasets and experimental configurations as regards setting GATree parameters.

There are two key theoretical directions that should be pursued. The most critical one is the emergence of the cost of space. As mentioned, while we can keep the space cost of indexing the instances to what we believe to be the theoretical minimum, $O(n)$, that cost is still a level of magnitude up from the previous situation. This trade-off must be researched and resolved. Actually, this direction of research is very closely related to the anytime learning flavor, where explicit quantification of such balancing decisions is factored into the top down induction of decision trees (Esmeir & Markovitch, 2007).

The time-space tradeoff has been usually addressed by the explicit consideration of caching intermediate tree evaluations. For example, Roberts (2003) factors the estimated speed up into every decision to cache and, eventually, suggests that any such decision must take into account the architectural details of the computer carrying out the computations, which is hardly a universal specification. We note that, while this approach has merits, it is in fact orthogonal to our approach; we do not deal with intermediate trees but with data that has to reside in the tree so that fitness can be reconstructed. Nevertheless, to factor in an estimate of when to apply one heuristic and to make that decision on the fly has been already demonstrated to be beneficial in incremental decision tree induction (Kalles & Papagelis, 2000).

A further direction concerns the broader applicability of indexing schemes for fitness functions calculated in a divide-and-conquer manner. In particular, we are interested to see whether there exist any applications that utilize functions which already incur the space cost that crept into the decision trees problem, yet have not been investigated regarding their potential for lossless fitness inheritance.

Last but not least, it is not trivial to investigate afresh the importance of the "losslessness" property, since by relaxing it one might be able to use fitness functions that are not tied to the actual instance description. The work of Rokach (2008) and Mansour & McAllester (2000) is along this direction; the latter specifically approach fitness calculations in the context of tree-subtree properties and their notion of "compositional" building of decision trees is related to our cross-over operator. The careful reader might question whether that takes us back to fitness approximation schemes, however when one frames the fitness question in terms of estimates and not in terms of actual accuracy calculations, it may be possible to deliver estimates in a divide-and-conquer approach. Even if approximation creeps back in, it is undoubtedly a step of progress to be able to accurately point out the assumptions and their impact.

Nevertheless, the top item in our agenda is of an experimental (and development) nature. The GATree system has been updated and savings in quite a few standard benchmarks have been easy to obtain, however that direction must effectively deal with implementation details to reduce the extent to which any savings are compromised by data structures housekeeping.

## 7. Acknowledgements

All work reported in this paper is research not previously undertaken or published by the author.

The opening of Section 2 borrows some text from Papagelis & Kalles (2001).

Stavros Zygounas (2004) first ventured into experimenting with GATree and speed-up techniques and has influenced the experimental understanding of what reasonable speed-up during evolution might be.

An anonymous reviewer pointed out that root crossover is very fast regardless of data structures and several reviewers have made numerous comments that substantially improved the presentation of this work and the proper acknowledgement of work by other researchers.